\def\year{2022}\relax
\documentclass[letterpaper]{article} % DO NOT CHANGE THIS
\usepackage{aaai22}  % DO NOT CHANGE THIS
\usepackage{times}  % DO NOT CHANGE THIS
\usepackage{helvet}  % DO NOT CHANGE THIS
\usepackage{courier}  % DO NOT CHANGE THIS
\usepackage[hyphens]{url}  % DO NOT CHANGE THIS
\usepackage{graphicx} % DO NOT CHANGE THIS
\usepackage{natbib}  % DO NOT CHANGE THIS AND DO NOT ADD ANY OPTIONS TO IT
\usepackage{caption} % DO NOT CHANGE THIS AND DO NOT ADD ANY OPTIONS TO IT
\DeclareCaptionStyle{ruled}{labelfont=normalfont,labelsep=colon,strut=off} % DO NOT CHANGE THIS
\usepackage{algorithm}
\usepackage{algorithmic}

\usepackage{newfloat}
\usepackage{listings}
\floatstyle{ruled}
\newfloat{listing}{tb}{lst}{}
\floatname{listing}{Listing}
\usepackage[utf8]{inputenc} % allow utf-8 input
\usepackage[T1]{fontenc}    % use 8-bit T1 fonts
\usepackage{hyperref}       % hyperlinks
\usepackage{url}            % simple URL typesetting
\usepackage{booktabs}       % professional-quality tables
\usepackage{amsmath,amsfonts,amsthm,amssymb}       % blackboard math symbols
\usepackage{nicefrac}       % compact symbols for 1/2, etc.
\usepackage{microtype}      % microtypography
\usepackage{xcolor}         % colors

\usepackage{graphicx, graphics, color}
\usepackage[normalem]{ulem}
\usepackage{caption,subcaption}
\usepackage{wrapfig}

\newtheorem{proposition}{Proposition}%[section]
\usepackage{bm} 
\usepackage{enumitem,kantlipsum}

\newcommand\x{\bm{x}}
\newcommand\fvec{\bm{f}}

\newcommand\bvec{\bm{b}}

\newcommand\W{\mathbf{W}}
\newcommand\y{\bm{y}}
\newcommand\z{\bm{z}}

\title{Understanding the Role of Self-Supervised Learning in Out-of-Distribution Detection Task}

\author {
    Jiuhai Chen ,\textsuperscript{\rm 1 $^{\dag}$}
    Chen Zhu, \textsuperscript{\rm 1}
    Bin Dai \textsuperscript{\rm 2}
}
\affiliations {
    \textsuperscript{\rm 1} University of Maryland\\
    \textsuperscript{\rm 2} Samsung Research China-Beijing\\
    \{jchen169, chenzhu\}@umd.edu, bin2.dai@samsung.com
}

\begin{document}

\maketitle
% \footnotetext[2]{Work done during internship at Amazon Web Services Shanghai AI Lab.}
\begin{abstract}
  Self-supervised learning (SSL) has achieved great success in a variety of computer vision tasks. However, the mechanism of how SSL works in these tasks remains a mystery. In this paper, we study how SSL can enhance the performance of the out-of-distribution (OOD) detection task. We first point out two general properties that a good OOD detector should have: 1) the overall feature space should be large and 2) the inlier feature space should be small. Then we demonstrate that SSL can indeed increase the intrinsic dimension of the overall feature space. In the meantime, SSL even has the potential to shrink the inlier feature space. As a result, there will be more space spared for the outliers, making OOD detection much easier. The conditions when SSL can shrink the inlier feature space is also discussed and validated. By understanding the role of SSL in the OOD detection task, our study can provide a guideline for designing better OOD detection algorithms. Moreover, this work can also shed light to other tasks where SSL can improve the performance.
\end{abstract}

\section{Introduction}
Self-supervised learning (SSL)~\citep{doersch2015unsupervised} has attracted much attention recently because it is very useful in a broad range of computer vision tasks including representation learning~\citep{kolesnikov2019revisiting,doersch2015unsupervised}, supervised learning~\citep{lee2019rethinking}, semi-supervised learning~\citep{tran2019semi} and out-of-distribution (OOD) detection~\citep{hendrycks2019using}. Two major streams of SSL algorithms have been proposed so far. The first stream creates a pretext learning task such as predicting the context~\citep{doersch2015unsupervised}, image rotation~\citep{gidaris2018unsupervised} and recovering positions of the permuted image patches~\citep{noroozi2016unsupervised}. The proxy task is used as either a standalone task to learn the feature representations or an auxiliary task to enhance the performance of the major task. Another stream of works is contrastive learning~\citep{chen2020simple,chen2020improved}. Instead of trying to predict the proxy label, these algorithms encourage the features of the transformed images to be close to each other.

Though a lot of progress has been made to make SSL achieve better performance on multiple tasks, the understanding of the mechanism of SSL is quite limited. In this paper, we study a specific task, namely the OOD detection task, and try to understand how SSL works in this task. We argue that a good OOD detection model should have a big overall feature space but small inlier feature space. As a result, there will be more space spared for outliers, making the OOD detection task much easier. We then show that SSL can indeed increase the intrinsic dimension of the overall feature space thanking to the auxiliary head of the proxy task. Meanwhile, we discover that SSL can even shrink the inlier feature space under mild conditions even that the overall feature space is expanded. This property further enhances the performance of OOD detection. Even when the conditions are violated such that SSL fails to shrink the inlier feature space, it will hardly enlarge it because all the inliers receive the same supervision from the SSL head. The inlier feature space will keep the same. Since the overall feature space is expanded, the OOD performance will also be improved.

Though our analysis only focuses on the OOD detection task, it can also shed light to other tasks where SSL can work. The fact that SSL can enlarge the overall feature space and sometimes shrink the inlier feature space is probably one of the key mechanisms that SSL can work in a variety of tasks. 

The contribution of this paper is two-fold:
\begin{enumerate}[wide, labelwidth=!, labelindent=2pt]
    \item We explicitly point out that a good OOD detector should have a large overall feature space but small inlier feature space. Though this argument is not surprising, it provides basis for understanding how SSL works in the OOD detection task and also guidelines for designing better OOD detection models.
    \item For the first time, we solidly unveil the mechanism of SSL in the OOD detection task. The model trained with SSL agrees extremely well with the properties mentioned above.
    On one hand, we demonstrate that SSL can increase the intrinsic dimension of the overall feature space both theoretically and empirically. On the other hand, SSL can shrink the inlier feature space under mild conditions. By providing more spared space for the outliers, SSL enhances the OOD detection performance. Further discussion about when SSL can shrink the inlier feature space provides a guideline about how to select the transformations in OOD detection. Even when the conditions for SSL to shrink the inlier feature space are violated, we show that SSL will hardly increase the inlier feature space though the overall feature space is expanded. %The explanation about how SSL works in the OOD detection also shed lights to other tasks that use SSL.
    %\item Connections between SSL and other OOD detection algorithms like outlier exposure and contrastive learning is also discussed, providing a more thorough understanding of different algorithms.
\end{enumerate}

The rest of the paper is organized as follows. We first discuss the related work about both SSL and OOD detection in Section~\ref{sec:related_work}. After introducing the preliminary and discussing what a good OOD detection model should be like in Section~\ref{sec:preliminary}, we show that SSL can enlarge the feature space in Section~\ref{sec:overall_feature} both theoretically and empirically. In Section~\ref{sec:inlier_feature}, we show that SSL can shrink the inlier feature space under mild conditions. The conditions when SSL can reduce the inlier feature space is also discussed followed by a conclusion in Section~\ref{sec:discussion}.

\section{Related Work}
\label{sec:related_work}

\subsection{Self-supervised learning}
SSL is proposed for better learning the visual representations in computer vision tasks. Early works mainly focus on exploring different pretext tasks. Autoencoder~\citep{ackley1985learning} can be regarded as the very first work of SSL which uses pixel values as the pretext task. Other proxy tasks are proposed since then~\citep{doersch2015unsupervised, dosovitskiy2015discriminative, gidaris2018unsupervised, larsson2016learning}. Recent progress on self-supervised learning has demonstrated the effectiveness of contrastive learning in various domains~\citep{chen2020simple, he2020momentum, oord2018representation, caron2020unsupervised, srinivas2020curl}.

Besides using SSL to learn visual representations for downstreaming tasks~\citep{kolesnikov2019revisiting,dosovitskiy2015discriminative,noroozi2016unsupervised}, SSL can also be applied to many other tasks to enhance the performance. For example, \citet{lee2019rethinking} incorporates SSL to a supervised learning framework and achieves higher classification accuracy. \citet{tran2019semi} uses SSL to improve the semi-supervised learning performance. \citet{hendrycks2019using} demonstrates that SSL can improve model robustness and greatly benefit OOD detection. Beyond this, SSL is also proven to be very useful in natural language processing~\citep{mikolov2013efficient,lan2019albert} and speech recognition~\citep{baevski2020wav2vec}.

Despite SSL has witnessed great empirical success across multiple domains, the understanding of how SSL works is rare. \citet{wang2020understanding} attempts to understand contrastive representation learning through alignment and uniformity. \citet{tian2021understanding} proposes a novel theoretical framework to understand SSL methods with dual pairs of deep ReLU networks.

\subsection{Out-of-Distribution Detection}
OOD detection has been a long history. Most existing works are in supervised representations~\citep{liang2017enhancing,  chalapathy2018anomaly, lee2018simple, hendrycks2016baseline, hendrycks2019deep}. These algorithms train a model that produces a score indicating how likely the input sample is an inlier. For example, \citet{hendrycks2016baseline} utilizes probabilities from softmax distributions as anomaly scores to detect outliers. Recently, many explorations have been with self-supervised learning~\citep{hendrycks2019using, tack2020csi, sehwag2021ssd}. \citet{hendrycks2019using} trains a classifier with an auxiliary self-supervised rotation loss. \citet{tack2020csi} proposes contrasting shifted instance (CSI), inspired by the framework of contrastive learning of visual representations.

\section{Preliminary}
\label{sec:preliminary}
In this section, we first introduce the notations and formulate the problem of multi-class OOD detection. Then we generally discuss what a good OOD detection model should be like. 

\subsection{Multi-Class OOD Detection Formulation}
Let $(\x,y)$ be the data pair that follows the groundtruth data distribution $\mathcal{D}$. Here $\x\in\mathbb{R}^D$ is a data point in a $D$ dimensional input space and $y$ is an integer ranges from $1$ to $C$, where $C$ is the number of classes. A deep neural network with parameter $\bold{\theta}$ transforms a point $\x$ in the image space to a feature $\fvec(\x;\bold{\theta})$ in the $\kappa$-dimensional feature space. We omit the arguments in $\fvec(\x;\bold{\theta})$ for convenience hereafter if there is no confusion. The feature is then fed into a linear classifier with parameter $\W\in\mathbb{R}^{\kappa\times C}$ and outputs a multinomial distribution $p_\bold{\theta}(y|\x)$ defined by $\text{softmax}(\W \fvec)$. We omit the bias $\bvec$ in the linear classifier to avoid the undue clutter since its role is minor.
%takes $\x$ as the input and produces a $K$ dimensional output vector $\hat{\y}_\bold{\theta}(\x)$ corresponding a multinomial distribution over the $K$ classes. The last layer of the deep network is usually a softmax layer such that $\hat{\y}_\bold{\theta}(\x)$ stands for a valid multinomial distribution.

A multi-class classification model is trained by using the expectation of cross entropy loss between the groundtruth $y$ and the output of the network $p_\bold{\theta}(y|\x)$ over the whole dataset, \emph{i.e.}
\begin{equation}
    L_{\text{vanilla}}(\bold{\theta}) = \mathbb{E}_{(\x,y)\sim\mathcal{D}} \left[ L_{CE}\left( \text{one\_hot}(y), p_\bold{\theta}(y|\x) \right) \right],
    \label{eqn:loss_ce}
\end{equation}
where $L_{CE}(p,q)$ stands for the cross entropy between the distributions $p$ and $q$.

After minimizing the objective (\ref{eqn:loss_ce}) w.r.t. $\bold{\theta}$, we can define a score based on $p_\bold{\theta}(y|\x)$ to determine whether the input $\x$ is an inlier from the data distribution $\mathcal{D}$ or an irrelevant outlier. Intuitively speaking, if $\x$ is an inlier, the network will be very confident about which class $\x$ belongs to. Thus there will exist a dimension $c$ such that $p_\bold{\theta}(y=c|\x)$ is very close to $1$. On the other hand, the network will be confused if $\x$ is an outlier, causing that $p_\bold{\theta}(y=c|\x)$ is relatively small for all possible $c$.

Based on this intuition, a straight-forward design of score is $\max_c p_\bold{\theta}(y=c|\x)$, which is adopted in \citet{hendrycks2019using}. Another scheme is to calculate the KL divergence between $p_\bold{\theta}(y=c|\x)$ and a uniform distribution over all the classes. Of course there are other score designs that can achieve even better results. However this is not our focus herein. In all our experiments, we will use $\max_c p_\bold{\theta}(y=c|\x)$ as the score.

There are three commonly adopted evaluation metrics for OOD detection: 1) area under the receiver operating characteristic curve (AUROC)~\citep{davis2006relationship}, 2) area under the precision-recall curve (AUPR)~\citep{manning1999foundations} and 3) false positive rate at $N\%$ true positive rate (FPR)~\citep{hendrycks2019deep}. Both AUROC and AUPR measure the performance of the detector across various thresholds. Higher value means better performance. FPR measures the performance at a strict threshold. In our experiments, we fix the threshold to be $95\%$. For FPR, lower value means better performance.
%For evaluating the performance of OOD detection, area under the receiver operating characteristic curve (AUROC) is the most commonly adopted metric~\citep{davis2006relationship}. A greater AUROC is better, and an uninformative detector has an AUROC of $50\%$. Area under the precision-recall curve (AUPR) is useful when anomalous samples are infrequent \citep{manning1999foundations}. AUROC and AUPR are two criteria measure the performance of detector across various thresholds, the false positive rate at $N\%$ true postive rate (FPRN) measures the performance at one strict threshold \citep{hendrycks2019deep}.

Self-supervised learning creates a proxy task by introducing a set of transformations $\{R_t\}_{t=1}^T$. A typical choice of $R_t$ is a rotation transformation of $t\times 90^\circ$, where $t\in\{0, 1, 2, 3\}$. The proxy task takes $R_t(\x)$ as the input and aims to predict the transformation type $t$. It shares the same feature network as the major task but uses a different output head. Specifically, it has another linear classifier with parameter $\W_{p}\in\mathbb{R}^{\kappa\times T}$. This linear classifier outputs a multinomial distribution $p_\bold{\theta}(t^\prime|R_t(\x))$ defined by $\text{softmax}(\W_p \fvec(R_t(\x);\bold{\theta}))$. The objective for the proxy task is the cross entropy between $p_\bold{\theta}(t^\prime|R_t(\x))$ and the groundtruth transformation $t$, \emph{i.e.}
\begin{equation}
    L_{\text{SSL}}(\bold{\theta}) = \mathbb{E}_{\x\sim\mathcal{D}} \left[\frac{1}{T} \sum_{t=1}^T L_{CE}\left( \text{one\_hot}(t), p_\bold{\theta}(t^\prime|R_t(\x) \right) \right].
\end{equation}
The final objective function then becomes $L_\text{vanilla}+\lambda L_\text{SSL}$ with $\lambda$ being a hyper parameter tuning the weights of the objectives of the major task and the proxy task. The proxy task is only used in the training phase. In the test phase, we only use the head of the major task.

\subsection{Conditions of A Good OOD Detection Model}

Denote the outlier data distribution as $\mathcal{D}_o$, which is usually unknown in the training phase\footnote{Of course we can introduce some outlier datasets into the training phase based on our prior knowledge. This method is known as outlier exposure \citep{hendrycks2019deep}. However, it is impossible to cover all outlier modes even with outlier exposure.}. The model projects both inlier and outlier into the feature space. Let $\mathcal{F}_{\text{inlier}}$ and $\mathcal{F}_{\text{outlier}}$ be the feature sets of the inlier and the outlier respectively. They both occupy some space in the feature space. A good OOD detection model should be able to separate the inlier set and the outlier set very well. A quantitative measurement of how well $\mathcal{F}_{\text{inlier}}$ and $\mathcal{F}_{\text{outlier}}$ separates each other is the intersection over union (IoU).
%defined as
%\begin{equation}
%    \text{IoU} = \frac{\mu(\mathcal{F}_{\text{inlier}} \cap \mathcal{F}_{\text{outlier}})}{\mu(\mathcal{F}_{\text{inlier}} \cup \mathcal{F}_{\text{outlier}})},
%\end{equation}
%where $\mu$ is a kind of measure that assigns a value to any set in the feature space $\mathbb{R}^D$. Based on this definition, we wish the model to produce large union and small intersection. However, in practice, we do not know the outlier distribution $\mathcal{D}_o$ such that it is impossible to obtain $\mathcal{F}_{\text{outlier}}$ given the feature network. 
So for a good OOD detection model, we wish the intersection between $\mathcal{F}_\text{inlier}$ and $\mathcal{F}_\text{outlier}$ to be small while the union of them to be large.

Unfortunately we do not know the distribution of outlier $\mathcal{D}_o$ in practice. To compromise, we replace the outlier feature set with the overall feature set $\mathcal{F}=\{ f(\x) | \x \in [0,1]^d \}$. Here we assume that the input $\x$ is an image and the pixel values are normalized to range $[0, 1]$. The feature of any possible outlier is covered by the overall feature set. It is straight forward to generalize the definition of the overall feature set to any other type of input data. In this case $\mathcal{F}_\text{inlier}$ becomes a subset of $\mathcal{F}$. Such a replacement is reasonable in the sense that the outlier may be anywhere in the input space. Note that it is also very likely that there exist some outliers such that the corresponding feature belongs to $\mathcal{F}_\text{inlier}$. These samples are known as adversarial samples.

Since $\mathcal{F}_\text{inlier}$ is a subset of $\mathcal{F}$, the intersection of $\mathcal{F}_\text{inlier}$ and $\mathcal{F}$ is $\mathcal{F}_\text{inlier}$ and the union of them is $\mathcal{F}$. With this compromise, we can conclude two necessary conditions of a good OOD detection model: 1) It should have a large overall feature set and 2) It should produce a small inlier feature set.

These two necessary conditions provide us a high-level intuition. We then need a specific metric to measure the size of the feature set and study the property of the models produced by SSL with this specific metric. In this paper, we use the intrinsic feature dimension to measure the feature space size. Though the dimension of the feature space ($\kappa$) is often very large, the features usually lie on a low-dimensional subspace embedded in this huge space. The dimension of the subspace is called the intrinsic dimension. It equals to the rank of the feature matrix. We can use this value to measure the size of the feature set. In practice, it is almost impossible for the features to exactly lie on a low-dimensional subspace. To compromise, we can calculate the singular values of the feature matrix and then normalize them by the maximum singular value. The curve of the normalized singular values in descending order can depict the feature space in a more subtle way. Throughout the paper, we use this curve to describe the size of different feature sets. A small feature space should have a curve that descends faster than a large feature space.

It needs to be pointed out that there are many other evaluation metrics to measure the size of different feature sets. We did not claim that our choice is the best one. Our main purpose herein is to show how SSL can change the property of the feature spaces via a specific metric. The high-level intuition is still likely to hold with other metrics.

\section{SSL Expands the Overall Feature Space}
\label{sec:overall_feature}
As we have mentioned, the dimension of the feature space $\kappa$ is usually very large but the features actually lie in an extremely low dimensional subspace. We study the intrinsic dimension of the overall feature space in this section.

Consider a deep neural network with the following diagram 
\begin{equation}
\x \xrightarrow {\text{Non-linear}}\z \xrightarrow{\W_f}\fvec\xrightarrow{\W}\bm{\eta}\xrightarrow{\textnormal{Softmax}}\bm{\hat{y}}.
\label{eqn:diagram}
\end{equation}
The input $\x$ is first projected to a $M$-dimensional feature map $\z$ via a nonlinear deep neural network. The feature vector $\fvec$ is obtained by $\W_f\z$ where $\W_f$ is a $\kappa\times M$ trainable matrix. The feature vector $\fvec$ is then transformed to the logit $\bm{\eta}$, which then produces the output $\hat{\y}$ by applying softmax operation. Such a network structure is very similar to a feed-forward convolutional network, where $\z$ is the second last feature map and $\fvec$ is the final feature vector. 
%In some cases, $\fvec$ is obtained by first applying a convolution operation on $\z$ and then globally averaging pooling the feature map. This is equivalent to first applying a linear transformation on $\z$ such that the spatial size of the feature map becomes the same as the convolutional kernel size and then linearly transform the feature map to $\fvec$ via the kernel. 
The only difference is that (\ref{eqn:diagram}) does not have a nonlinear activation after $\fvec$.We then have the following proposition about the overall feature space:

\begin{proposition}
Assume there is a network with structure shown in (\ref{eqn:diagram}). The objective function is $L_\text{vanilla}$. The parameters are optimized using the stochastic gradient descent algorithm with weight decay. Suppose $\W_f$ and $\W$ converge to $\W_f^*$ and $\W^*$ respectively, then we have
\iffalse
\begin{equation}
    \text{span}[\W_{f,1\cdot}^*, \W_{f,2\cdot}^*, ..., \W_{f, M\cdot}^*] = \text{span}[\W_{\cdot 1}^*, \W_{\cdot 2}^*, ..., \W_{\cdot C}^*],
    \label{eqn:span}
\end{equation}
\fi

\begin{align}
    &\text{span}[\W_{f,1\cdot}^*, \W_{f,2\cdot}^*, ..., \W_{f, M\cdot}^*] \nonumber \\
    =~~ &\text{span}[\W_{\cdot 1}^*, \W_{\cdot 2}^*, ..., \W_{\cdot C}^*],
    \label{eqn:span}
\end{align}

where $\W_{f,m\cdot}^*$ is the $m$-th row of $\W_f^*$ while $\W_{\cdot c}^*$ is the $c$-th column of $\W^*$. In addition, $\text{span}[\cdot]$ means the subspace spanned by the basis in the arguments.
\label{prop:span}
\end{proposition}

The proof of the proposition can be found in the supplementary material. 
Both $\W_{f,m\cdot}^*$ and $\W_{\cdot c}^*$ are $\kappa$-dimensional vectors in the feature space. Note that $M$, which is the dimension of the hidden state $\z$, is usually the same as $\kappa$ while $C$ is often much smaller. This proposition suggests that $\{\W_{f,m\cdot}^*\}_{m=1}^M$ only spans an extremely low dimensional space despite $M$ is very large. For any input $\x$, its feature is a linear combination of $\{\W_{f,m\cdot}^*\}_{m=1}^M$, meaning that it lies in a very small subspace embedded in the huge $\kappa$-dimensional feature space. 

Proposition~\ref{prop:span} offers an upper bound of the rank of the overall feature space. When a vanilla method is adopted, the dimension of the feature space is bounded by the number of classes $C$. Then we consider adding SSL regularization in the diagram (\ref{eqn:diagram}). After obtaining $\fvec$, there is a new head for the proxy task. The trainable parameter in this head is $\W_p$ which is a $\kappa\times T$ matrix. Then we have 

\begin{proposition}
Assume there is a new head with $T$ outputs in the network shown in (\ref{eqn:diagram}). The objective function is $L_\text{vanilla}+\lambda L_\text{SSL}$ with $\lambda > 0$. The parameters are optimized using the stochastic gradient descent algorithm with weight decay. Suppose $\W_f$, $\W$ and $\W_p$ converge to $\W_f^*$, $\W^*$ and $\W_p^*$ respectively, then we have

\begin{align}
    &\text{span}[\W_{f,1\cdot}^*, \W_{f,2\cdot}^*, ..., \W_{f, M\cdot}^*] \label{eqn:span_ssl} \\
    =~ &\text{span}[\W_{\cdot 1}^*, \W_{\cdot 2}^*, ..., \W_{\cdot C}^*, \W_{p,\cdot 1}^*, ..., \W_{p,\cdot T}^*],
    \nonumber
\end{align}

\label{prop:span_ssl}
\end{proposition}
Again the dimension of the overall feature space is bounded by a relatively small value. However, after introducing the proxy task, the upper bound of the feature space dimension increases. Moreover, considering that the major task and the proxy task are completely different, they are very likely to use different features for classification, leading to that the intersection between $\text{span}[\W_{\cdot 1}^*, \W_{\cdot 2}^*, ..., \W_{\cdot C}^*]$ and $\text{span}[\W_{p,\cdot 1}^*, \W_{p,\cdot 2}^*, ..., \W_{p,\cdot T}^*]$ is very small. Thus the proxy task will expand the overall feature dimension and leave more wiggle room for the potential outliers.

\subsection{Empirical Validation of the Feature Dimension}
\label{sec:feat_dim}
We empirically validate both proposition~\ref{prop:span} and \ref{prop:span_ssl} on two datasets: Cifar-10 \citep{krizhevsky2009learning} and SVHN~\citep{netzer2011reading}. For both datasets, we use two different methods: 1) vanilla and 2) SSL with rotation auxiliary loss (SSLR). There is no auxiliary SSL head for the vanilla method. The SSLR scheme rotates the image by $t\times90^\circ$ where $t\in\{0,1,2,3\}$ and the proxy task tries to predict the rotation angle. 
% The SSLT algorithm uses two proxy tasks. It randomly translates the image by 0, 1 or 2 pixels both vertically and horizontally. The first proxy task predicts the horizontal translation while the second proxy task predicts the vertical translation. So both proxy heads have $3$ output dimensions.

For Cifar-10, we use WideResNet40-2~\citep{zagoruyko2016wide} with dropout rate $0.3$. Here $40$ stands for the depth of the network while $2$ represents the widen factor of a base architecture. The hidden state after global pooling is $128$-dimensional. To make the feature network structure consistent with (\ref{eqn:diagram}), we add a linear layer after the global pooling layer to linearly transform the $128$-dimensional hidden state to a $128$-dimensional feature vector. For SVHN, we use VGG16~\citep{simonyan2014very} architecture. Since VGG16 already has two full connected layers after global pooling, we remove the activation function of the last fully connected layer and directly feed the feature without activation into the classifier. Note that the reason that we use different architectures for different datasets is to verify our proposition in different scenarios. The result will be similar if we use VGG16 for Cifar-10 and WideResNet for SVHN.
We use the momentum optimizer with Nesterov momentum being 0.9~\citep{sutskever2013importance}. The cosine learning rate schedule is adopted with the initial learning rate being 0.1. The batch size is 128 and we train the model for 200 epochs. 

\begin{figure}[t!]
     \centering
     \begin{subfigure}[b]{0.42\textwidth}
     \centering 
         \includegraphics[width=\linewidth]{ 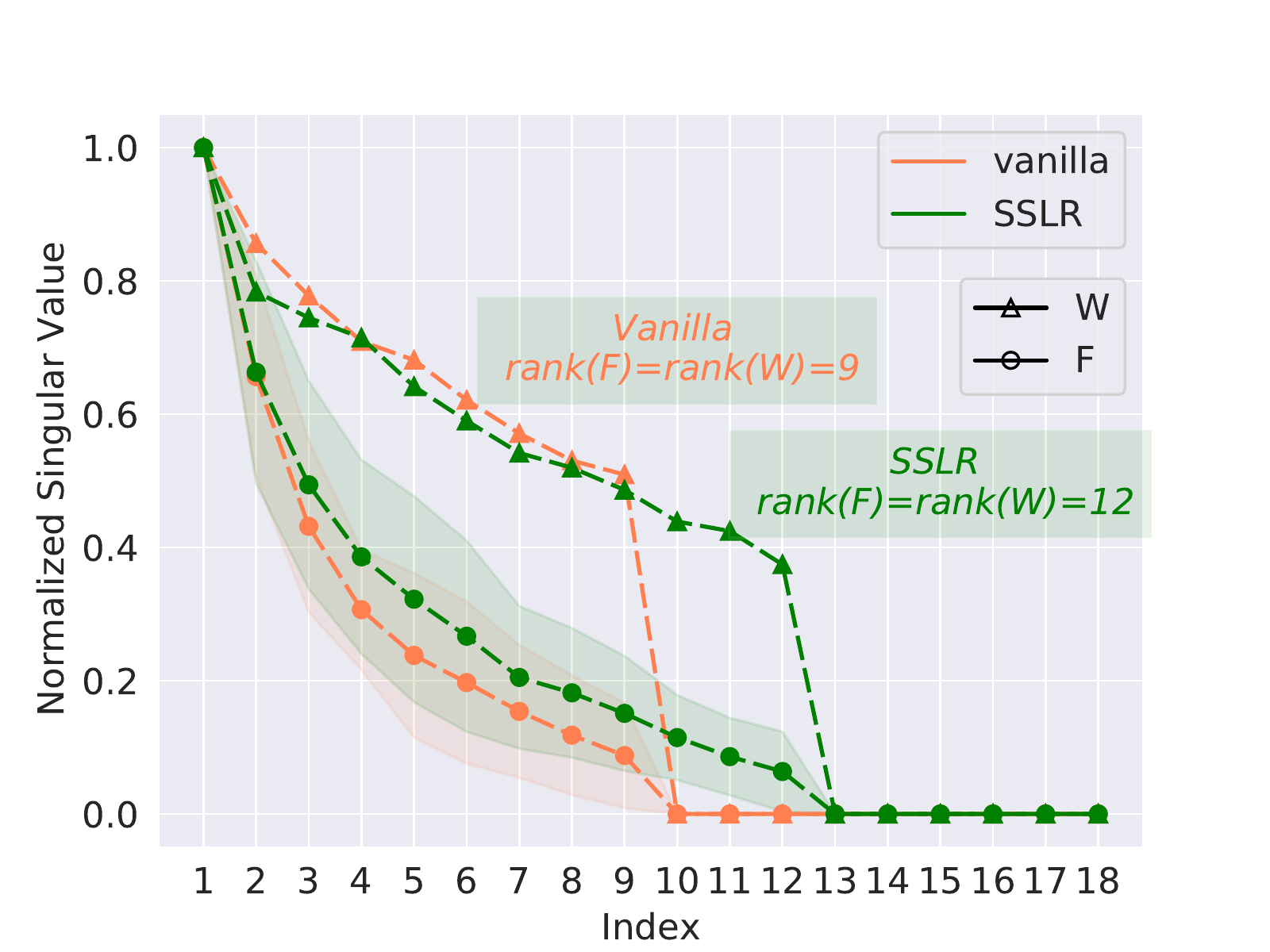}
         \caption{Inlier: Cifar-10 (WideResNet40-2).}
         \label{fig:rank_cifar10}
    \end{subfigure}
    \begin{subfigure}[b]{0.42\textwidth}
    \centering
        \includegraphics[width=\linewidth]{ 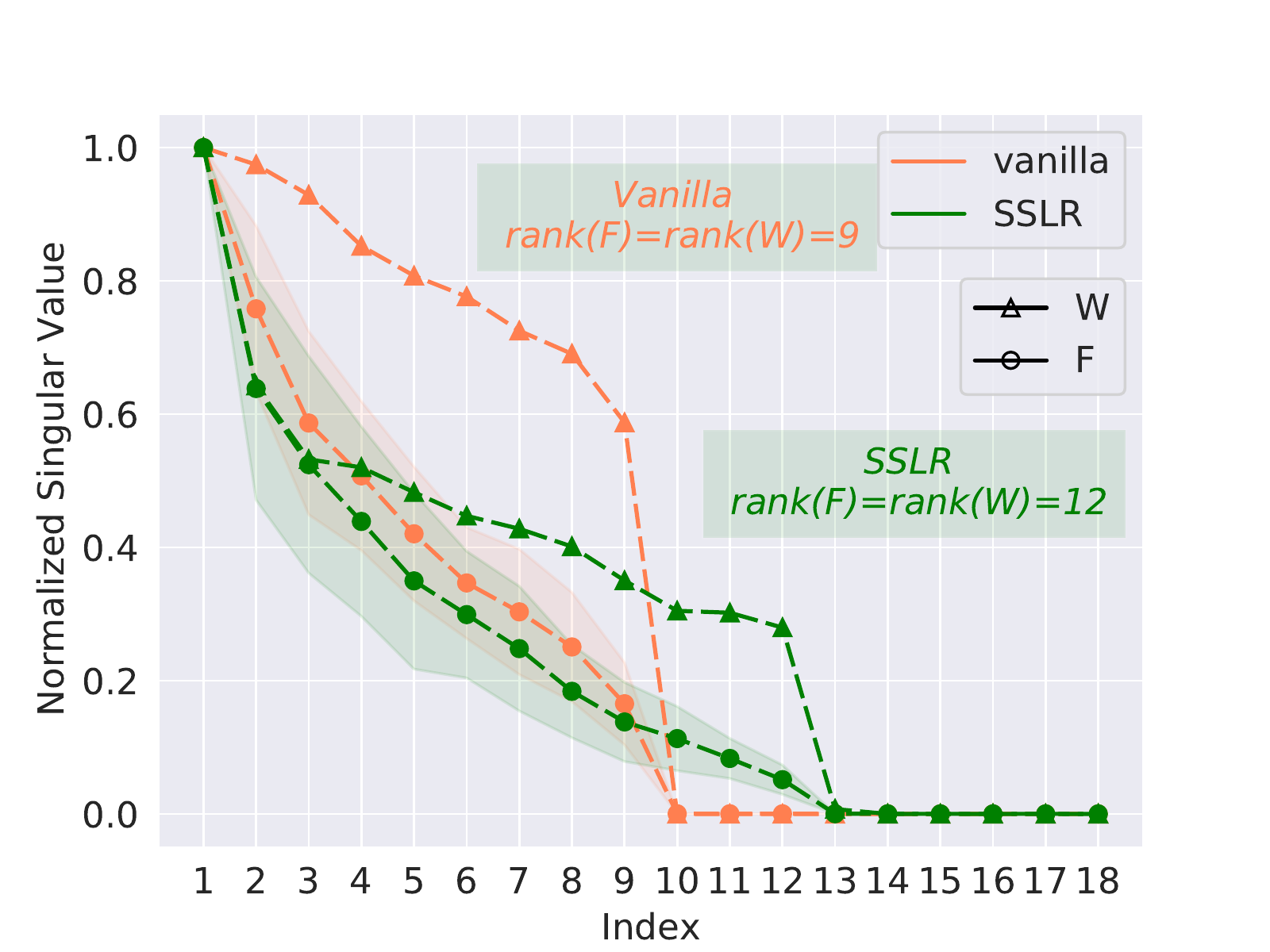}
        \caption{Inlier: SVHN (VGG16).}
        \label{fig:rank_svhn}
    \end{subfigure}
    \caption{Normalized singular value of the feature matrix and the classification weight matrix. The normalized singular values decrease to $0$ at exactly the same time for both matrices. Using SSL clearly increases the rank of both the weight and feature matrices.}
    \label{fig:rank}
\end{figure}

After training the network on the inlier dataset, we feed the network with both inliers and outliers to obtain the feature vectors to measure the intrinsic dimension of the overall feature space. The outliers include Cifar-100~\citep{krizhevsky2009learning}, Textures~\citep{cimpoi14describing}, Place365~\citep{zhou2017places}, Gaussian (each dimension is sampled from an isotopic Gaussian distribution), Rademacher (each dimension is $-1$ or $1$ with equal probability sampled from a symmetric Rademacher distribution) and Blob image (Blobs data consist in algorithmically generated amorphous shapes with definite edges) \citep{hendrycks2019deep}. After obtaining these features, we calculate the singular values of the feature matrix and normalize them with the maximum singular value. Similarly, we concatenate the weight matrices in the heads of both major task and proxy task and then calculate the normalized singular values of the concatenated weight matrix. The normalized singular values of both feature matrix and weight matrix are ordered in descending order and shown in Figure~\ref{fig:rank}.

First of all, in both vanilla and SSLR scenarios, the feature rank and the weight rank exactly matches. The normalized singular values decreases to exactly $0$ at the same time, validating proposition~\ref{prop:span} and \ref{prop:span_ssl}. Secondly, we observe that the rank of feature matrix in the vanilla case is exactly $C-1$ (orange line, $C=10$). This is a reasonable result. The $c$-th output dimension corresponds to the direction $\W_{\cdot c}^*$. The loss tries to separate all these directions apart. Given the first $C-1$ directions, the optimal $C$-th direction should be close to the opposite direction of the mean direction of the first $C-1$ directions, making the rank less than $C$. After adding the proxy task, the rank of the features also increases, as the green line shows. The rotation proxy task has $4$ output dimensions, so the rank of the features is $(10-1)+(4-1)=12$. 

\subsection{Feature Dimension in a More Realistic Scenario}
\label{sec:feat_dim_real}
The experiments in Section~\ref{sec:feat_dim} agree very well with proposition~\ref{prop:span} and \ref{prop:span_ssl}. We also witness that the intrinsic dimension of the overall feature space increases when SSL is added. However, in a more realistic scenario, we directly use the output of the global pooling layer as the feature rather than linearly transform it to another feature space (In the VGG16 case, there will be a nonlinear activation after the fully connected layer in practice). The diagram of the deep network then becomes
\begin{equation}
\x \xrightarrow {\text{Non-linear}}\fvec \xrightarrow{\W}\bm{\eta}\xrightarrow{\textnormal{Softmax}}\bm{\hat{y}}.
\label{eqn:diagram2}
\end{equation}
In this case, the feature vector $\fvec$ only receives gradients in the directions $\{\W_{\cdot c}\}_{c=1}^C$. The gradients will propagate back to the nonlinear network to move $\fvec$ in the space spanned by $\{\W_{\cdot c}\}_{c=1}^C$ such that it can very well align with $\W_{\cdot c}$ of the corresponding class. On the other hand, The component of $\fvec$ that is orthogonal to $\{\W_{\cdot c}\}_{c=1}^C$ will be eliminated by weight decay. Note that though the weight decay is applied on the weights rather than the feature vectors, it still constraints the length of $\fvec$. Since the length of $\fvec$ is limited, its component orthogonal to $\{\W_{\cdot c}\}_{c=1}^C$ should be as small as possible to minimize the objective. So if the capacity of the network is big enough, $\fvec$ will lie in the subspace spanned by $\{\W_{\cdot c}\}_{c=1}^C$. 
%To be more formal, we have
%\begin{proposition}
%Assume there is a deep network shown in (\ref{eqn:diagram2}) with enough capacity. The feature vectors of the training data are denoted as $\{\fvec^{(i)}\}_{i=1}^N$ where $N$ is the total number of the training data. The feature vectors are bounded such that $||\fvec^{(i)}||_2\le A$. The network is optimized using objective (\ref{eqn:loss_ce}). If the network is globally optimized with $\W=\W^*$, then 
%\begin{equation}
%    \fvec^{(i)} \in \text{span}[\W_{\cdot 1}^*, \W_{\cdot 2}^*, ..., \W_{\cdot C}^*].
%\end{equation}
%\label{prop:span_real}
%\end{proposition}

\begin{figure}[t!]
     \centering
     \begin{subfigure}[b]{0.43\textwidth}
         \centering
         \includegraphics[width=\linewidth]{ 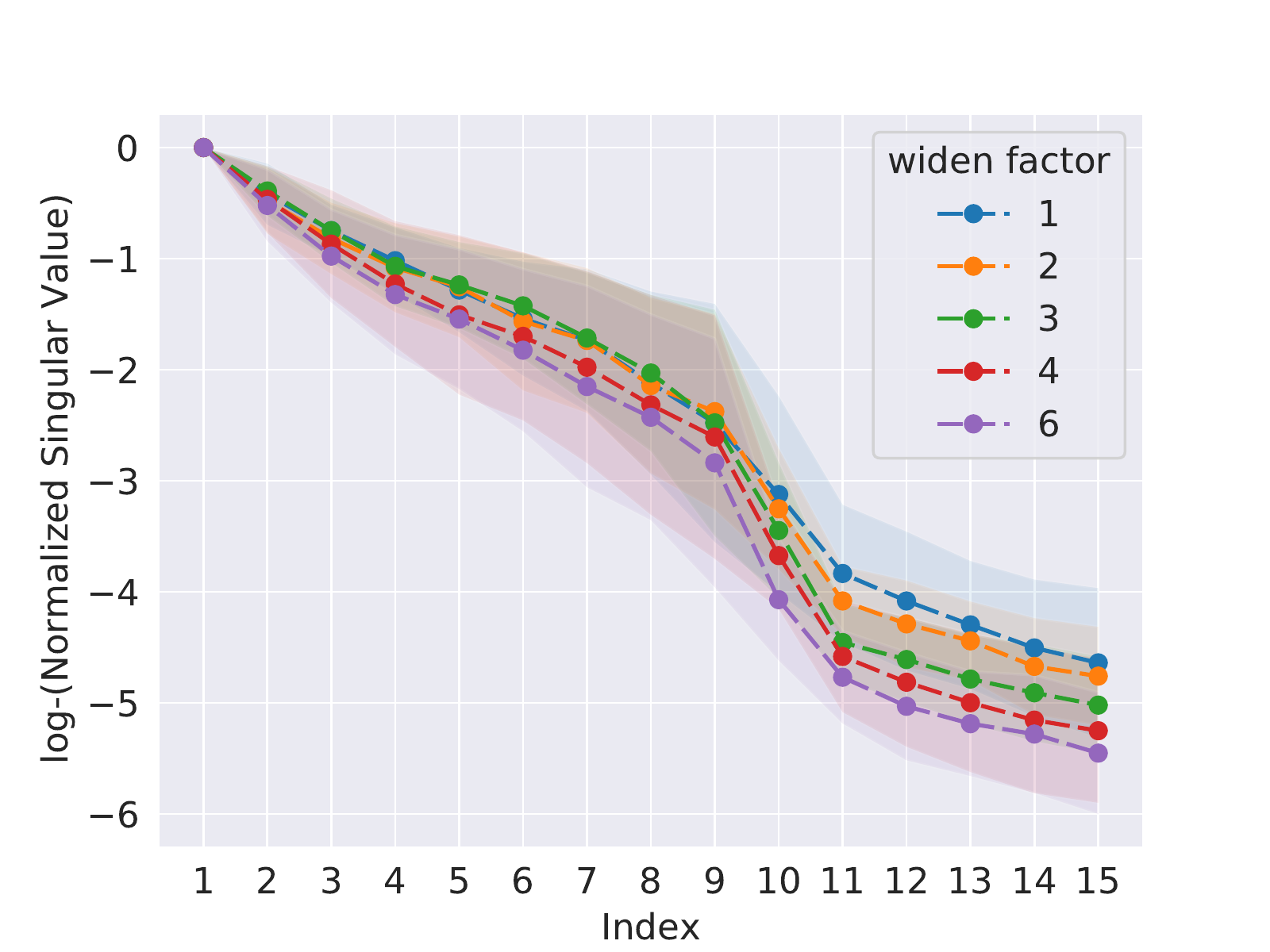}
         \caption{Without SSL.}
     \label{fig:widen_factor_vanilla}
     \end{subfigure}
     \hfill
     \begin{subfigure}[b]{0.43\textwidth}
         \centering
         \includegraphics[width=\linewidth]{ 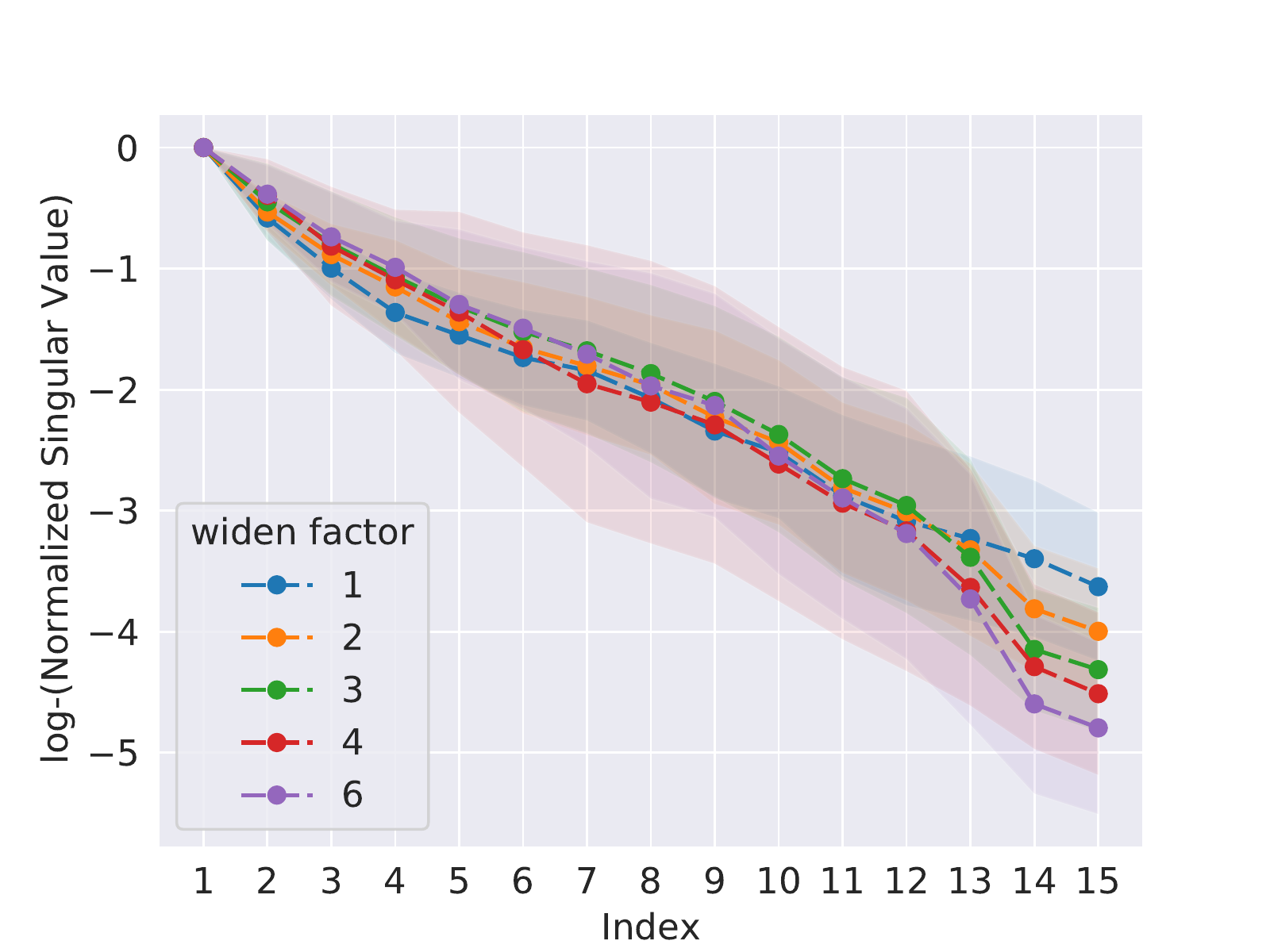}
         \caption{With SSL.}
     \label{fig:widen_factor_ssl}
     \end{subfigure}
     \caption{Logarithm of the normalized singular values of the feature matrix with different widen factors. As the widen factor (network capacity) increases, the pattern of the normalized singular values become closer to that in Figure~\ref{fig:rank}.}
     \label{fig:widen_factor}
\end{figure}

To validate the analysis above, we run the experiments with different network capacity. Again we use the WideResNet architecture but with different widen factors to control the network capacity. Also, we remove the linear transformation layer that is used in Section~\ref{sec:feat_dim}. The training scheme is exactly the same as that in Section~\ref{sec:feat_dim}. Similarly, we feed both the inliers and the outliers to the optimized network to obtain the features and then calculate the normalized singular values of the feature matrix. The logarithm of the normalized singular values of different widen factors are shown in Figure~\ref{fig:widen_factor}. 
In this more realistic case, the singular values do not decrease exactly to $0$ at the $10$th dimension in the vanilla case (Figure~\ref{fig:widen_factor_vanilla}). However, as the network capacity increases, the singular values decrease more sharply near the $10$th dimension, which is consistent with our analysis. When SSL is introduced, the normalized singular values decrease more slowly. They start to decrease very sharply near the $13$th dimension.

\begin{figure}[t!]
     \centering
     \begin{subfigure}[b]{0.4\textwidth}
         \centering
         \includegraphics[width=\textwidth]{ 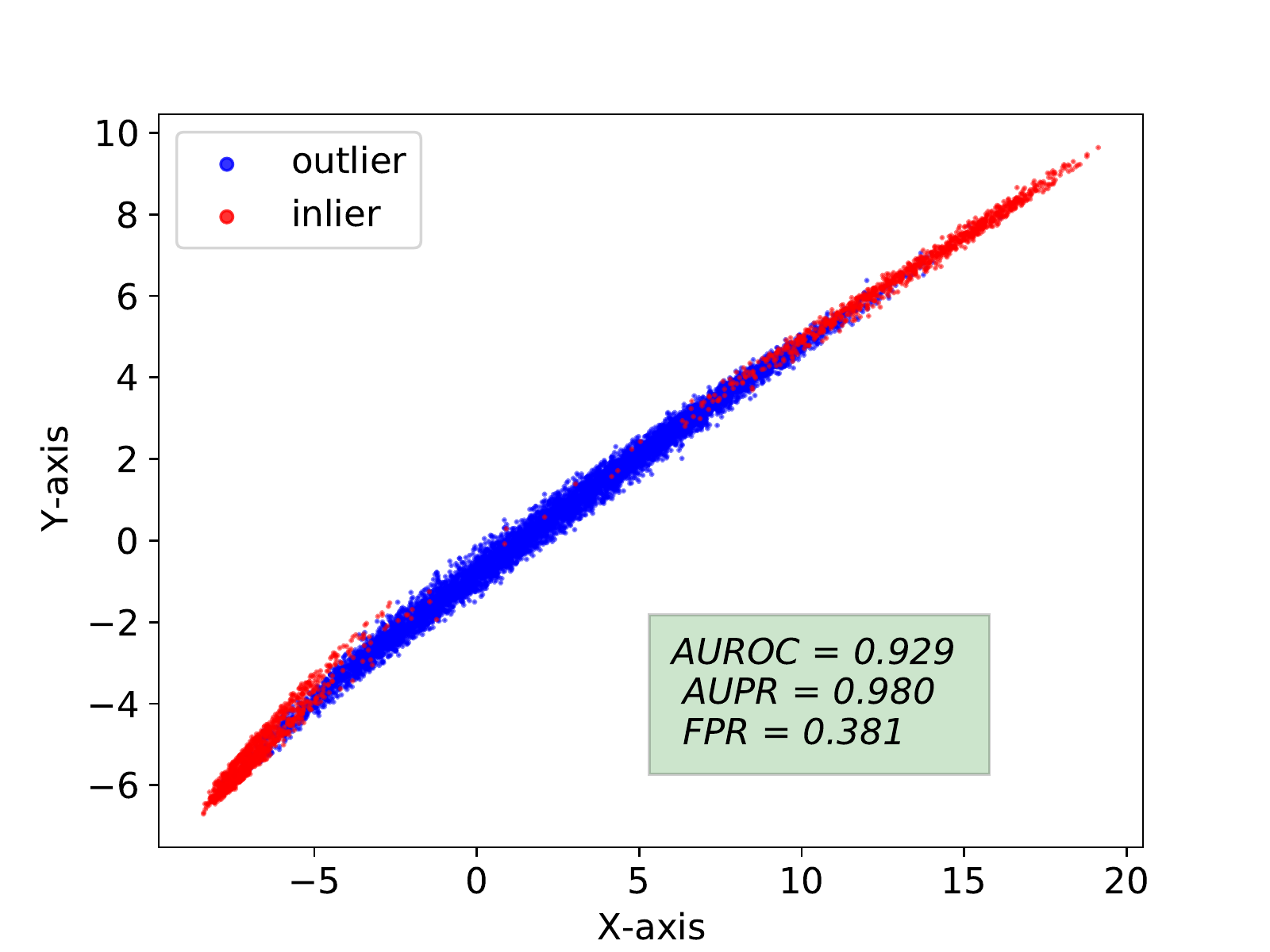}
         \caption{Without SSL.}
         \label{fig:mnist_vanilla}
     \end{subfigure}
     \hfill
     \begin{subfigure}[b]{0.4\textwidth}
         \centering
         \includegraphics[width=\textwidth]{ 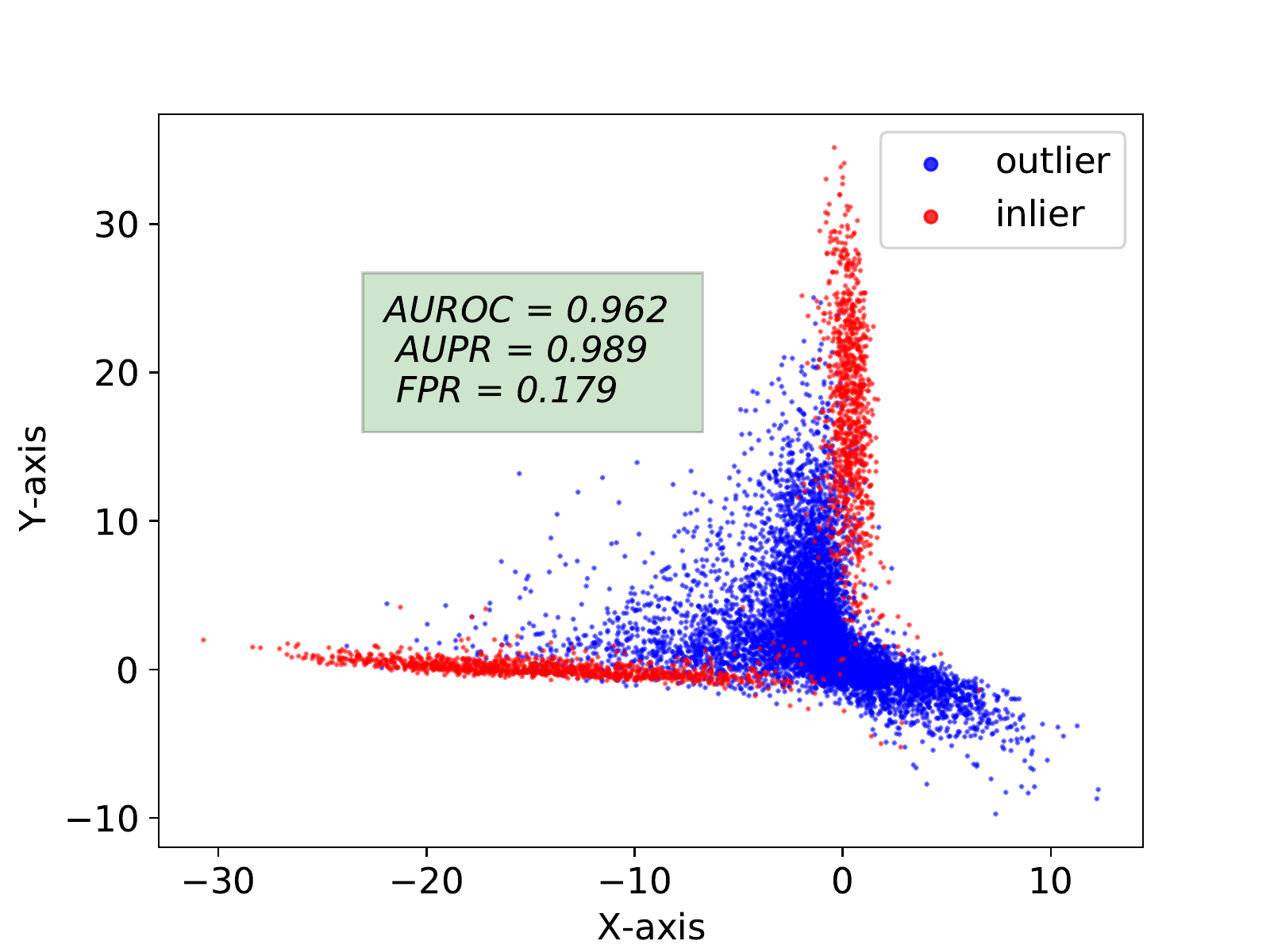}
         \caption{With SSL.}
         \label{fig:mnist_rot}
     \end{subfigure}
        \caption{
        Feature space without and with SSL. When there is no SSL, the features lie in an approximately $1$-dimensional subspace. Introducing SSL expands the intrinsic dimension of the overall feature space and keeps the intrinsic dimension of each inlier cluster unchanged.
        }
        \label{fig:mnist}
\end{figure}

To better visualize how SSL expands the overall feature dimension, we run a simple experiment on MNIST dataset. Only digit $0$ and $1$ are used as the inliers. The major task is a simple binary classification task. The other $8$ digits are regarded as outliers. For the proxy task, we only use $2$ kinds of transformations: rotation by $0^\circ$ and $180^\circ$. So the proxy task also becomes a binary classification task. We use the LeNet-5~\citep{lecun2015lenet} architecture but reduce the dimension of the feature layer from $84$ to $2$ for visualization convenience. For optimizer, we choose a cosine learning rate schedule with initial learning rate being 0.01 and Nesterov momentum being 0.9. The batch size is $128$ and we train the network for $10$ epochs. The tuning parameter $\lambda$ for rotation auxiliary is $4$. Figure~\ref{fig:mnist} shows the features of both inliers (red) and outliers (green). The dimension of the overall feature space clearly increases from $1$ to $2$ after the SSL head is used.

\section{When SSL Can/Cannot Shrink the Inlier Feature Space}
\label{sec:inlier_feature}

We have shown that SSL can expand the overall feature space in Section~\ref{sec:overall_feature} both theoretically and empirically. However, it does not necessarily mean that SSL can improve the OOD detection performance. SSL can only work when the inlier feature space does not significantly increase. Fortunately this is indeed the case. Moreover, we find that SSL can sometimes even shrink the inlier feature space though the overall feature space is expanded, making the OOD detection even easier.

% \begin{figure}[t!]
%     \centering
%     \includegraphics[width=0.5\linewidth]{OOD_Draft/Figures/cifar10-intra-class.pdf}
%     \caption{Mean and standard deviation of the normalized singular values of each inlier class of Cifar-10 trained on WideResNet40-2. The curves of SSLR and PSSLR descend even faster than that of vanilla. }
%     \label{fig:cifar10-intra-class}
% \end{figure}

\begin{figure}[t!]
     \centering
     \begin{subfigure}[b]{0.45\textwidth}
         \centering
         \includegraphics[width=\textwidth]{ 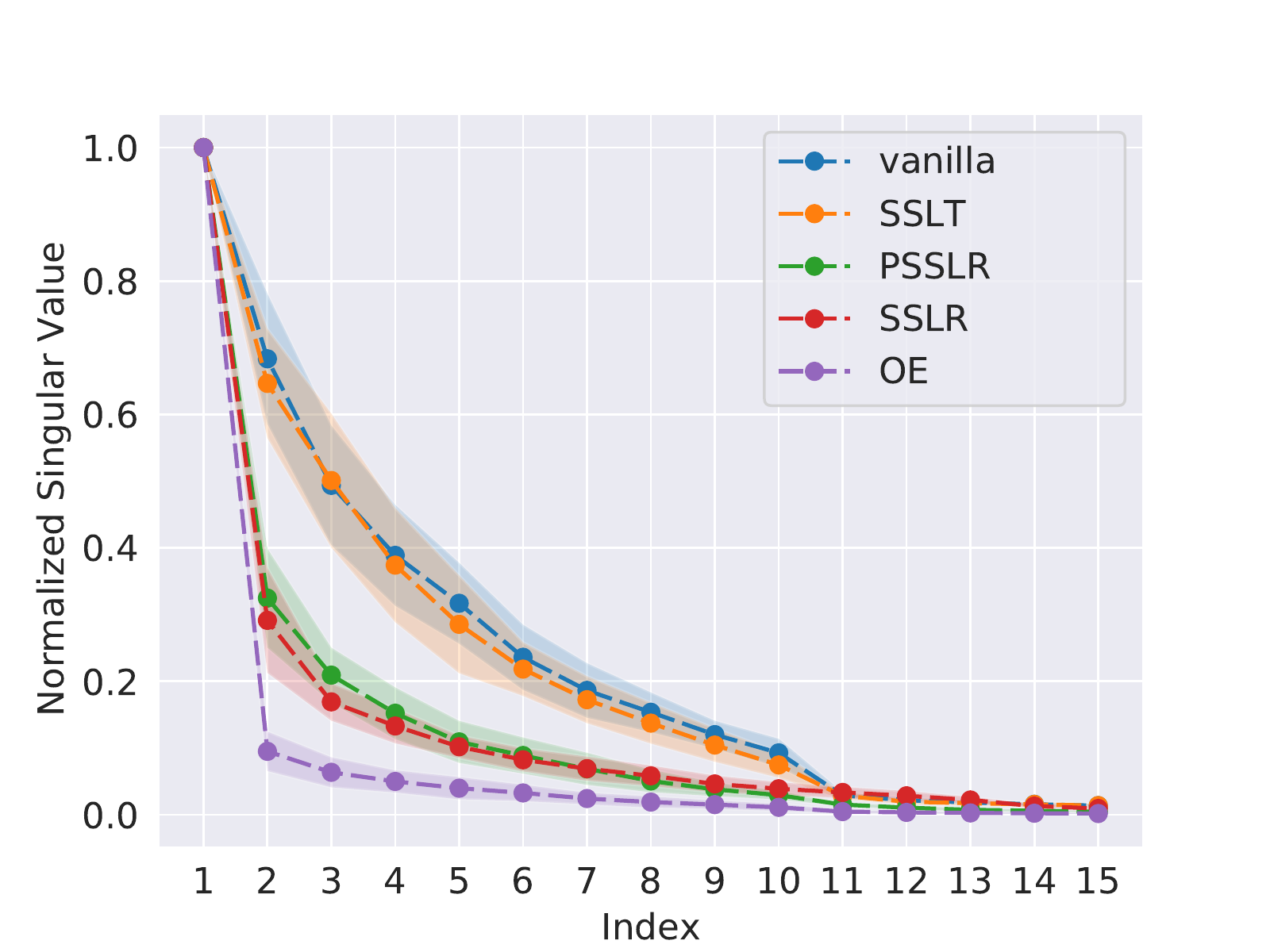}
         \caption{Cifar10 trained on WideResNet40-2}
         \label{fig:cifar10-intra-class}
     \end{subfigure}
     \hfill
     \begin{subfigure}[b]{0.45\textwidth}
         \centering
         \includegraphics[width=\textwidth]{ 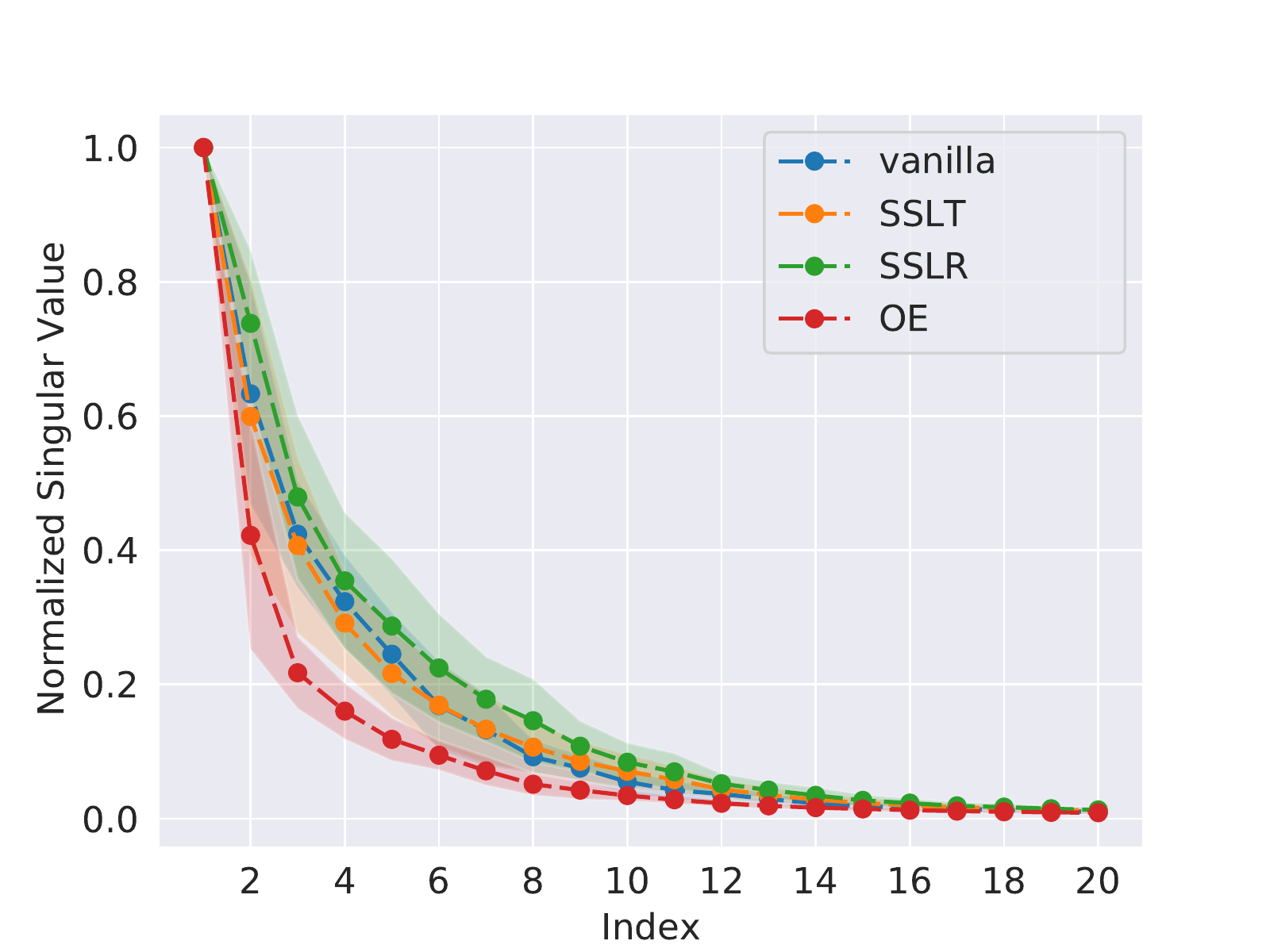}
         \caption{MNIST trained on Lenet5}
         \label{fig:mnist-intra-class}
     \end{subfigure}
        \caption{Mean and standard deviation of the normalized singular values of each inlier class.
        }
        % \label{fig:mnist}
\end{figure}

We first extract the inlier features of Cifar-10 dataset trained on WideResNet40-2 (corresponding to Figure~\ref{fig:widen_factor}). For each inlier class, we calculate the normalized singular values and sort them in descending order. The mean and standard deviations of the normalized singular values over all the classes are shown in Figure~\ref{fig:cifar10-intra-class}. It is obvious that the curve of SSLR (red) descends much faster than that of vanilla (blue), indicating that SSL encourages the inliers to use even a smaller feature space.

The reason that SSL shrinks the inlier feature space is that it has a regularization effect. When the network is wide enough, the features in every layer tend to use more dimensions than needed. If the transformed images have different low-level patches, the early convolution layers should leave space for the patches of the transformed images, encouraging the inlier patches to use a smaller space. As a result, the inlier will occupy a smaller feature space in all the layers, including the last feature layer.

This regularization effect can happen even without the SSL loss. We run another experiment with hyper-parameter $\lambda$ being $0$. We call this scheme pseudo SSLR (PSSLR for short). In this case, the transformed images only contributes to updating the running mean and running variance in the batch normalization layers. They do not have any corresponding loss term. Since the running mean and running variance count both the original images and the transformed images, it can also shrink the inlier features in all the layers. The curve is also shown in Figure~\ref{fig:cifar10-intra-class} (green line). Surprisingly, the curve is very close to the SSLR case. Its OOD detection performance is also much better than the vanilla scheme (We will show the detailed values later).

\subsection{Necessary Conditions for SSL to Shrink the Inlier Feature Space}
Unfortunately, this regularization effect does not always happen. There are two necessary conditions for it to work: 1) The low-level patches of the transformed images have significantly different statistics than that of the original images; 2) The network capacity is relatively large.
\iffalse
\begin{enumerate}
    \item The low-level patches of the transformed images have significantly different statistics than that of the original images.
    \item The network capacity is relatively large.
\end{enumerate}
\fi 

Based on our analysis above, it is obvious that the regularization effect will disappear if the low-level patches of the transformed images are close to those of the original images. To validate this, we run an experiment using SSL with translation transformations (SSLT). In SSLT, the images are vertically or horizontally translated by $-1, 0$ or $1$ pixel. The corresponding curve of the normalized singular values (orange curve in Figure~\ref{fig:cifar10-intra-class}) descends almost at the same rate as that of the vanilla case.

We then test the OOD detection performance of the models trained using vanilla, SSLT, PSSLR and SSLR schemes on multiple outlier datasets including Cifar-100, SVHN, Texture, LSUN, Place365, Gaussian, Rademacher and Blob. The average performance over all the outlier datasets is shown in Table~\ref{tab:cifar10}. The vanilla scheme produces the worst performance. The SSLR method, which can simultaneously expand the overall feature space and shrink the inlier feature space is the best among all four methods. SSLT can only expand the overall feature space while PSSLR can only shrink the inlier feature space. Their results are between vanilla and SSLR, which is reasonable.

\begin{table}
\centering
\begin{tabular}{@{\hskip1pt}c@{\hskip1pt}|@{\hskip1pt} c  c c@{\hskip1pt}}
    \hline
    & AUROC & AUPR&FPR  \\
    \hline
    Vanilla & 0.9068 &0.8384& 0.4376  \\
    SSLT & 0.9182 &0.8404& 0.3845 \\
    PSSLR & 0.9485 &0.8827  &0.2106  \\
    SSLR & 0.9652 &0.9080& 0.1599  \\
%    OE &0.97  &0.89& 0.11 \\
    \hline
\end{tabular}
\caption{The performance of outlier detector trained by different methods. Cifar 10 is inlier and all the evaluation metrics are mean value over different outliers (Cifar-100, SVHN, Texture, LSUN, Place365, Gaussian, Rademacher and Blob). }
\label{tab:cifar10}
\end{table}

Another necessary condition for SSL to shrink the inlier feature space is that the network capacity should be large enough. If the network is relatively thin, the feature space may be too small to capture all the patterns of the inlier patches. When transformed images with different low-level patches are added in the training phase, both the original image patches and the transformed image patches are entangled in the same low-dimensional space in the early layers, leaving the hard classification problem to future layers. This prohibits the last feature layer using a smaller inlier feature space to achieve good classification performance.

An example is shown in Figure~\ref{fig:mnist}, where the network architecture is LeNet5. It only has $6$ channels in the first convolution layer, which is obviously not sufficient to capture all the modes of $5\times5$ patches. Fortunately, even when the network capacity is sometimes not enough, SSL will not significantly increase the inlier feature space. In Figure~\ref{fig:mnist_rot}, the inliers still lie in an approximately $1$-dimensional subspace. We further validate this point on the full MNIST dataset using LeNet5 (with the last layer dimension being $84$ rather than $2$). Three different schemes are adopted: vanilla, SSLT and SSLR. The mean and standard deviation of the normalized singular values over all the inlier classes are shown in Figure~\ref{fig:mnist-intra-class}. The curve of SSLT decreases at the same rate as that of vanilla, meaning that the inlier feature sizes are very similar in both cases. This is reasonable since they have the same statistics of the low-level patches. In SSLR, where the transformed images become significantly different, the curve is only slightly higher than the vanilla case, indicating that the inlier feature space does not significantly increase.

%For Figure~\ref{fig:mnist-intra-class}, we also run experiments on Cifar-10 and SVHN, the phenomenon is very similar. In addition, for Figure~\ref{fig:cifar10-intra-class}, experiments on MNIST and SVHN also have similar behaviors. This further demonstrates our arguments that SSL will not significantly expand the inlier feature space and sometimes can shrink it.

\subsection{Connection with Outlier Exposure}
Conventional wisdom thinks that SSL serves as a kind of outlier exposure (OE) \citep{hendrycks2019deep} in OOD detection. Based on our analysis, this is not the whole story. OE will shrink the inlier feature space by explicitly introducing outliers into the training phase. We add a line corresponding to OE in both Figure~\ref{fig:cifar10-intra-class} and \ref{fig:mnist-intra-class}. It is obvious that the normalized singular values of OE decreases at the fastest rate.
However, there is no mechanism that it can expand the overall feature space. As a result, OE performs no better than SSL though it has even smaller inlier feature space.

\section{Discussion}
\label{sec:discussion}
In this paper, we try to understand the mechanism of SSL in the OOD detection task. We first explicitly pointed out that a good OOD detection model should have large overall feature space and small inlier feature space. Then we demonstrate that SSL can indeed expand the overall feature space both theoretically and empirically. Moreover, SSL can also shrink the inlier feature space under mild conditions. The necessary conditions about when SSL can shrink the inlier feature space is presented and empirically validated. Even when the necessary conditions are violated, SSL is still able to keep the inlier feature space unchanged. The conditions of a good OOD detection model we presented provide us a guideline to design better OOD detectors, which will be our future work.

%It need to be emphasized that the feature space intrinsic dimension and the feature space volume used in this paper are not the only metrics to measure the size of the feature space. There could exist other kinds of measurements that are even better. We will briefly talk about this in the supplementary file. The high level intuition provides a guideline for us to design better OOD detectors, which will be our future work.

Though this work only studies the mechanism of SSL in OOD detection task, it can also shed light on other tasks where SSL works. For example, in supervised learning and semi-supervised learning, SSL expands the feature space such that it can disentangle the class information and the rotation (or other transformations) information. In this way, it can avoid a certain kind of overfitting and improve the classification performance. In addition, the regularization mechanism we showed in Section~\ref{sec:inlier_feature} probably also plays a role in other tasks. Of course the mechanism analyzed in this paper is not the whole story about how SSL works in all the tasks, but we believe this work is a meaningful step towards better understanding the SSL mechanism.

% References and End of Paper
% These lines must be placed at the end of your paper




\vskip 0.3in
\bibstyle{aaai22}
\bibliography{VI}

\end{document}